\documentclass[11pt,a4paper]{article}
\usepackage[hyperref]{emnlp2018}
\usepackage{times}
\usepackage{latexsym}
\usepackage{graphicx}
\usepackage{float}
\usepackage{url}

\usepackage{supertabular}

\usepackage{array}
\newcolumntype{L}[1]{>{\raggedright\let\newline\\\arraybackslash\hspace{0pt}}m{#1}}
\newcolumntype{C}[1]{>{\centering\let\newline\\\arraybackslash\hspace{0pt}}m{#1}}
\newcolumntype{R}[1]{>{\raggedleft\let\newline\\\arraybackslash\hspace{0pt}}m{#1}}

\aclfinalcopy % Uncomment this line for the final submission
\def\aclpaperid{282} %  Enter the acl Paper ID here

\title{Learning a Policy for Opportunistic Active Learning}

\author{Aishwarya Padmakumar, Peter Stone and Raymond J. Mooney \\
  Department of Computer Science \\
  University of Texas at Austin \\
  {\tt \{aish,pstone,mooney\}@cs.utexas.edu}
  }

\date{}

\begin{document}
\maketitle
\begin{abstract}
Active learning identifies data points to label that are expected to be the most useful in improving a supervised model.
Opportunistic active learning incorporates active learning into interactive tasks that constrain possible queries during interactions. 
Prior work has shown that opportunistic active learning can be used to improve grounding of natural language descriptions in an interactive object retrieval task.
%Prior work on opportunistic active learning has used a heuristic strategy to demonstrate the efficacy of opportunistic active learning. 
% In this work, we learn a policy for such an object retrieval task that trades off between completing the current interaction, and improving its model either for current or future interactions. 
In this work, we use reinforcement learning for such an object retrieval task, to learn a policy that effectively trades off task completion with model improvement that would benefit future tasks. 
\end{abstract}

\section{Introduction}

% What is active learning
In machine learning tasks where obtaining labeled examples is expensive, active learning is used to lower the cost of annotation without sacrificing model performance. 
Active learning allows a learner to iteratively query for labels of unlabeled data points that are expected to maximally improve the existing model. 
It has been used in a number of natural language processing tasks such as text categorization~\cite{lewis:sigir1994}, semantic parsing~\cite{thompson:ml99} and information extraction~\cite{settles:emnlp2008}.

The most commonly used framework for active learning is pool-based active learning, where the learner has access to the entire pool of unlabeled data at once, and can iteratively query for examples. 
In contrast, sequential active learning is a framework in which unlabeled examples are presented to the learner in a stream~\cite{lewis:sigir1994}.
For every example, the learner can decide whether to query for its label or not. 
This results in an additional challenge -- since the learner cannot compare all unlabeled data points before choosing queries, each query must be chosen based on local information only. 

% Multilabel learning
%Most active learning settings assume there is a single label for a data point. 
Multilabel active learning is the application of active learning in scenarios where multiple labels, that are not necessarily mutually exclusive, are associated with a data point~\cite{brinker:2006}. 
These setups often suffer from sparsity, both in the number of labels that are positive for a data point, and in the number of positive data points per label. 

% What is opportunistic active learning
Opportunistic active learning incorporates a form of multilabel sequential active learning into an interactive task. 
It was recently introduced for the task of interpreting natural-language object descriptions, motivated by the task of instructing a robot to retrieve a specific item ~\cite{thomason:corl17}.   
In this task, a human describes one of a set of objects in unrestricted natural language and the agent must determine which object was described. 
The agent is allowed to ask questions about other objects in the current environment to obtain labels that allow it to learn classifiers for concepts used in such descriptions.
As the questions are restricted to the objects available in the current interaction, the learning process across interactions can be seen as a form of multilabel sequential active learning. 
Further, the agent can either restrict itself to querying labels relevant to understanding the current description, or be opportunistic and query labels that can only aid future interactions -- for example querying whether some object is ``round'' when trying to understand the description ``a red box''.

More generally, in opportunistic active learning, an agent is engaged in a series of sequential decision-making tasks.
The agent uses one or more supervised models to complete each task.
Each task involves some sampled examples from a given feature space, and the agent is allowed to query for labels of these examples to improve its models for current and future tasks.
Queries in this setting have a higher cost than in traditional active learning as the agent may choose to query for labels that are not relevant for the current task, but expected to be of use for future tasks. Such 
opportunistic queries enable an agent to learn from a greater number of interactions, by allowing it to ask queries that would aid future tasks when it is sufficiently confident of completing the current task. 
They also allow an agent to focus on concepts that could have more impact than those relevant to the current task -- for example by choosing a frequently used concept as opposed to a rare one. 
Further, identifying which queries are optimal for model improvement is more difficult as the agent does not have access to the entire pool of unlabeled examples at any given time, similar to sequential active learning settings. 

Another sample application of opportunistic active learning could be in a task oriented dialog system providing restaurant recommendations to a user. In this case, a possible opportunistic query would be to ask the user for a Chinese restaurant they liked, when the user is searching for an Italian one. The query is not relevant to the immediate task of recommending an Italian restaurant but would improve the underlying recommendation system. 

Prior work on using opportunistic active learning in understanding natural-language object descriptions has shown that an agent following an opportunistic policy, that queries for labels not necessarily relevant to the current interaction, learns to perform better at identifying objects correctly over time~\cite{thomason:corl17}. 
However, this work only compares static policies that select actions based on manually-engineered heuristics. 
In this work, we focus on {\it learning} an optimal policy for this task using reinforcement learning, in the spirit of other recent attempts to learn policies for different types of active learning ~\cite{fang:emnlp2017,woodward:arxiv2017}. This allows an agent to choose whether or not to be opportunistic based on the specific interaction as well as the overall statistics of the dataset.  

Our learned policy outperforms a static baseline by improving its success rate on object retrieval while asking fewer questions on average. 
The learned policy also learns to distribute queries more uniformly across concepts than the baseline. 

\section{Related Work}

Active learning methods aim to identify examples that are likely to be the most useful in improving a supervised model. A number of metrics have been proposed to evaluate examples, including uncertainty sampling~\cite{lewis:sigir1994}, density-weighted methods~\citep{settles:emnlp2008}, expected error reduction~\cite{roy:icml2001}, query by committee~\cite{seung:colt1992}, and the presence of conflicting evidence~\citep{sharma:kdd2017}; as surveyed by \newcite{settles:2010}. 
Some of these metrics can be extended to the multilabel setting, by assuming that one-vs-all classifiers are learned for each label, and that all the learned classifiers are comparable~\cite{brinker:2006,singh:icaic2009,li:icip2004}. 
Label statistics have also been incorporated into heuristics for selecting instances to be queried~\cite{yang:kdd2009,li:ijcai2013}.  
There have also been Bayesian approaches that select both an instance and label to be queried~\cite{qi:pami2009,vasisht:kdd2014}. 
Our work aims to learn a policy for choosing between queries that can use information from many such indicators, but learns to combine them appropriately for a given task. 

\newcite{thomason:corl17} define the setting of opportunistic active learning, and apply it to an interactive task of grounding natural language descriptions of objects. They compare two static policies to demonstrate that using opportunistic queries improves task performance. 
We try to learn the optimal policy for this task using reinforcement learning, and compare to a policy similar to theirs. 

Recently, there has been interest in using reinforcement learning to learn a policy for active learning. 
\newcite{fang:emnlp2017} use deep Q-learning to acquire a policy that sequentially examines unlabeled examples and decides whether or not to query for their labels; using it to improve named entity recognition in low resource languages. 
Also, \newcite{bachman:arxiv2017} use meta-learning to jointly learn a data selection heuristic, data representation and prediction function for a distribution of related tasks. 
They apply this to one shot recognition of characters from different languages, and in recommender systems.
In contrast to these works, we learn a policy for a task that contains {\it both} possible actions that are active learning queries, {\it and} actions that complete the current task, thus resulting in a greater exploration-exploitation trade-off. 

More similar to our setup is that of \newcite{woodward:arxiv2017} which uses reinforcement learning with a recurrent-neural-network-based Q-function in a sequential one-shot learning task to decide between predicting a label and acquiring the true label at a cost. 
This setup also has a higher cost than standard active learning where the test set is separated out.
This is a continuous task without clearly separated interactions or episodes.
In our setting, each episode or interaction allows for querying and requires completion of an interaction, which further increases the trade-off between model improvement and exploitation.
Further, we consider a multilabel setting, which increases the number of actions at each decision step.

There are other works that employ various types of turn-taking interaction to learn models for language grounding. 
Some of these use a restricted vocabulary~\cite{cakmak:2010,kulick:ijcai2013}, or additional knowledge of predicates (for example that ``red'' is a color)~\cite{mohan:acs12}.
Others do not use active learning~\cite{kollar:rss13,parde:ijcai15,de:cvpr2017,yu:arxiv2017}, or do not learn a policy that guides the interaction~\cite{vogel:aaai10,thomason:ijcai16,thomason:corl17}. 

% Remove this if space is needed
Also related to our work is the use of reinforcement learning in dialog tasks, such as slot-filling and recommendation~\cite{wen:emnlp15,pietquin:tslp11}, understanding natural language instructions or commands~\cite{padmakumar:eacl17,misra:arxiv2017}, and open domain conversation~\cite{serban:aaai2016,das:cvpr2017}.
These typically do not use active learning.
In our task, the policy needs to trade-off model improvement against task completion.

\section{Opportunistic Active Learning}

Opportunistic Active Learning (OAL) is a setting that incorporates active learning queries into interactive tasks.
Let $O = \{o_1, o_2, \ldots o_n\}$ be a set of examples, and $M = \{m_1, m_2, \ldots m_k\}$ be supervised models trained for different concepts, using these examples. 
For the problem of understanding natural-language object descriptions, $O$ corresponds to the set of objects, $M$ corresponds to the set of possible concepts that can be used to describe the objects, for example their categories (such as \textit{ball} or \textit{bottle}) or perceptual properties (such as \textit{red} or \textit{tall}).  

In each interaction, an agent is presented with some subset $O_A \subseteq O$, and must make a decision based on some subset of the models $M_A \subseteq M$.
Given a set of objects $O_A$ and a natural language description $l$, $M_A$ would be the set of classifiers corresponding to perceptual predicates present in $l$. The decision made by the agent is a guess about which object is being described by $l$. 
The agent receives a score or reward based on this decision, and needs to maximize expected reward across a series of such interactions. 
In the task of object retrieval, this is a 0/1 value indicating whether the guess was correct, and the agent needs to maximize the average guess success rate. 

During the interaction, the agent may also query for the label of any of the examples present in the interaction $o \in O_A$, for any model $m \in M$. 
The agent is said to be opportunistic when it chooses to query for a label $m \notin M_A$, as this label will not affect the decision made in the current interaction, and
can only help with future interactions.  For example, given a description ``\textit{the red box}'', asking whether an object is \textit{red}, could help the agent make a better guess, but asking whether an object is \textit{round}, would be an opportunistic query.
Queries have a cost, and hence the agent needs to trade-off the number of queries with the success at guessing across interactions.

The agent participates in a sequence of such interactions, and the models improve from labels acquired over multiple interactions. Thus the agent's expected reward per interaction is expected to improve as more interactions are completed. 

This setting differs from the traditional application of active learning in the following key ways:
\begin{itemize}
\setlength\itemsep{0em}
\item The agent cannot query for the label of any example from the unlabeled pool. It is restricted to the set of objects available in the current interaction, $O_A$. 
\item The agent is evaluated on the reward per interaction, rather than the final accuracy of the models in $M$.
\item The agent may make opportunistic queries (for models $m \notin M_A$) that are not relevant to the current task.
\end{itemize}

Due to these differences, this setting provides challenges not seen in most active learning scenarios:
\begin{itemize}
\setlength\itemsep{0em}
\item Since the agent never sees the entire pool of unlabeled examples, it can neither choose queries that are globally optimal, nor use variance reduction strategies that still use near-optimal queries (such as sampling from a beam of near globally optimal queries). 
\item Since the agent is evaluated on task completion, it must learn to trade-off finishing the task with querying to improve the models.
\item The agent needs to estimate the usefulness of a model across multiple interactions, to identify good opportunistic queries.
\end{itemize}

\section{Task Setup}
\label{sec:task_setup}

We consider an interactive task where an agent tries to learn to ground natural-language object descriptions. 
Grounded language understanding is the process of mapping natural-language referring expressions to object referents in the world~\cite{thomason:ijcai16}.
We consider a grounded-language problem based on object retrieval -- given a free form natural-language description of an object, the agent needs to identify which of a set of objects is best described by the phrase~\cite{thomason:ijcai16,guadarrama:rss14}.
In this work, objects are presented as images, but the methods are applicable to any feature representation of objects.
We consider a task of interactive object retrieval where the agent is given a natural-language object description, and allowed to interact with the user before it attempts to guess the object being referred to.

In each interaction, the agent is presented with two sets of objects. 
The first set of objects is called the active training set, and is to be used by the agent to improve its model of object properties. 
The second set of objects is called the active test set, and the agent will have to retrieve an object from this set. 
The agent is provided with a natural language description of the object it is expected to retrieve.  

Before guessing, the agent is allowed to ask queries of the following two types:
\begin{itemize}
\setlength\itemsep{0em}
\item Label queries - A yes/no question about whether a predicate can be used to describe one of the objects in the active training set, e.g. ``Is this object yellow?''.
\item Example queries - Asking for an object, in the available training set, that can be described by a particular predicate, e.g. ``Show me a white object in this set.''. This is used for acquiring positive examples since most predicates tend to be sparse.~\footnote{Alternately, we could return all positive examples for the predicate in the active training set, but we chose to return a single example to allow the agent to minimize the amount of supervision obtained}
\end{itemize}

\begin{figure}[h]
\centering
\includegraphics[width=0.5\textwidth]{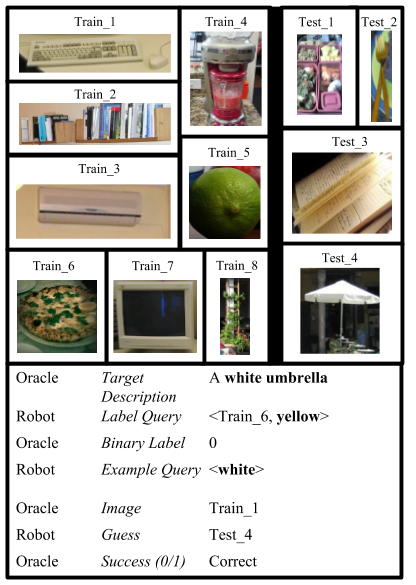}
\caption{A sample OAL interaction. Perceptual predicates are marked in bold.}
\label{fig:example}
\end{figure}

\noindent A sample interaction is shown in Figure \ref{fig:example}. 
The agent goes through a series of such interactions, and needs to learn to maximize the number of correct guesses across interactions, without frustrating the user with too many queries.
The separate active training set and active test set ensures that the agent needs to learn models for object descriptors. If queries and guessing were performed on the same set of objects, the agent could simply query whether each specific object satisfies each predicate in the description, and use this to guess. 

%Prior work has shown that for this task, an agent that makes opportunistic active learning queries learns to better understand object descriptions, than a more task oriented agent that restricts queries to those useful for understanding the description given in the current interaction~\cite{thomason:corl17}.
%However, this work only compares two possible policies that an agent performing this task could follow. 
%It does not attempt to learn an optimal policy for this task. 
%In this work, we model this interaction as a decision process, and attempt to learn an optimal policy for this task using reinforcement learning. 

In our experiments, we simulate such dialogs using the Visual Genome dataset~\cite{krishna:ijcv2017}; which contains images with regions (crops) annotated with natural-language descriptions. Bounding boxes of objects present in the image are also annotated, along with attributes of objects. Region descriptions, objects and attributes are annotated using unrestricted natural language, which leads to a diverse set of predicates. Using the annotations, we can associate a list of objects and attributes relevant to each image region, and use these to answer queries from the agent. 

For each interaction, we uniformly sample 4 regions to form the active test set, and 8 regions to form the active training set.~\footnote{The regions in the dataset are divided into separate pools from which the active training and active test sets are sampled (described as classifier-training and classifier-test sets in section \ref{ssec:sampling}), to ensure that the agent needs to learn classifiers that generalize across objects.}
One region is then uniformly sampled from the active test set to be the target object.
Its description, from annotations in the Visual Genome dataset, is provided to the agent to be grounded. 
The objects and attributes associated with active training regions are used to answer queries. 
A predicate is labeled as being applicable to a region if it is present in the list of objects and attributes associated with the region. 
In the rest of the paper, we use the terms object, image, and region interchangeably. 

\section{Methodology}

\subsection{Perceptual Predicates and Classifiers}
\label{ssec:perceptual_classifiers}

We assume that the description provided is a conjunction of one-word predicates. Given a description, the agent tokenizes it and removes stopwords. Each remaining word is stemmed and treated as a perceptual predicate. 
This method allows the agent to learn an open vocabulary of predicates, but unable to handle multi-word predicates or non-compositional phrases.

The agent learns a separate binary classifier for each predicate, and 
we represent images with a ``deep'' feature representation obtained from the penultimate layer of the VGG network~\cite{simonyan:arxiv2014} pretrained on ImageNet~\cite{russakovsky:ijcv2015}. 
The agent has no initial classifiers for any predicate, and learns these classifiers purely from labels acquired during interactions.   

\subsection{Grounding Descriptions}
\label{ssec:grounding}

The learned perceptual classifiers are used to ground natural language descriptions as follows.  Let $p_1, p_2, \ldots p_k$ be the predicates obtained from the natural language description. 
Let $d(p_i, o) \in \{-1, 1\}$ be the decision from the classifier for predicate $p_i$ for object $o$, and $C(p_i)$ be the estimated F1 of the classifier for $p_i$.~\footnote{F1 is estimated by cross-validation on the labels acquired for the predicate.}
Then the best guess, from the objects present, is chosen using the weighted sum of the decisions of the classifiers, using their estimated F1 as a weight:
$$
o_{guess} = \mbox{argmax}_{o \in O_A}{\sum_{i=1}^k{d(p_i, o) * C(p_i)}}
$$

\subsection{MDP Formulation}

We model the task as a Markov Decision Process (MDP).
An MDP is a tuple $\langle S, A, T, R, \gamma \rangle$, where $S$ is a set of states, $A$ is a set of actions, $T$ is a transition function, $R$ is a reward function and $\gamma$ is a discount factor. 
Each interaction is an episode in the MDP.
At any point, the agent is in a state $s \in S$, in our case consisting of the VGG features of the images in the current interaction, the predicates in the current description, and the agent's classifiers.  
The agent can choose from among actions in $A$, which include an action for guessing, and an action for each possible query the agent can make, including both label and example queries. 
The guess action always terminates the episode, and query actions transition the agent to a state $s' \in S$ as one of the classifiers gets updated. 
The agent gets a reward for each action taken. Query actions have a small negative reward, and guessing results is a large positive reward when the guess is correct, and a large negative reward when the guess is incorrect. 
In our experiments, we treat the reward values as hyperparameters that can be tuned. The best results were obtained with a reward of 200 for a correct guess, -100 for an incorrect guess and -1 for each query. 

\subsection{Identifying Candidate Queries}

In any interaction, the agent can make label or example queries. 
In a label query, the agent can ask for the label of any object for a specific predicate. If $O_A$ is the set of objects present in the active training set of the current interaction, and $P$ is the set of predicates that have been seen by the agent in all interactions so far, then the set of possible label queries is $P \times O_A$.  
Once the agent chooses a predicate $p$ and object $o$ to be queried, it obtains the corresponding label and can update its classifier for $p$.  In an example query, the agent asks for a positive example for any predicate $p \in P$. 
The agent will either receive a positive label for $p$ for some object $o \in O_A$ or learn that the label is negative $\forall\; o \in O_A$, and can appropriately update the classifier for $p$. 

Since $\vert P \vert$ grows across interactions as the agent encounters more predicates in descriptions, the number of candidate actions in a state increases over time, so
searching the entire space of possible queries can become intractable.
Hence, we identify a beam of promising queries that are then provided as candidate actions for the policy to choose among. 
Uncertainty sampling is a common method in pool-based active learning to identify the best example to improve a classifier. 
For a given predicate $p$, we use this to choose the best label query involving that predicate, picking that object $o \in O_A$ which is closest to the hyperplane of the classifier for $p$.  

However, it is more challenging to narrow down the number of predicates. 
\newcite{thomason:corl17} assume that an estimate of classifier accuracy is available, which is comparable across classifiers. They sample predicates with a probability inversely proportional to the estimated accuracy of the classifier. 
However, if the space of possible predicates is large, then this results in no classifier obtaining a reasonable number of training examples. 
In this scenario, it is desirable to focus on a small number of predicates, possibly stopping the improvement on a predicate once the classifier for it has been sufficiently improved. 
We sample queries from a distribution designed to capture this intuition. The probability assigned to a predicate by this distribution increases linearly, for estimated F1 below a threshold, and decreases linearly thereafter.~\footnote{The equation for this distribution with some further discussion on its design is included in the supplementary material.}
The number of queries sampled is a hyperparameter. We obtain the best results by sampling 3 queries of each type.
% We sample example queries uniformly at random. 

\subsection{Baseline Static Policy}
\label{ssec:static_policy}

As a baseline, we use a static policy similar to that used by \newcite{thomason:corl17}. At each state, a single label query and example query are sampled. The agent asks a fixed number of queries before guessing. \newcite{thomason:corl17} use thresholds that prevent queries from being asked when there are no predicates whose classifiers have sufficiently low estimated accuracy. Since we used a dataset with a much larger number of predicates, these thresholds were always crossed if the agent had even one candidate query.   

\subsection{Policy Learning}

We use the REINFORCE algorithm~\cite{williams:1992} to learn a policy for the MDP. 
The agent learns a policy $\pi(a \vert s; \theta)$, parameterized with weights $\theta$ that computes the probability of taking action $a$ in state $s$. Given a feature representation $f(s, a)$ for a state-action pair $(s, a)$, the policy is of the form:
$$
\pi(a \vert s; \theta) = \frac{e^{\theta^T f(s, a)}}{\sum_{a'}e^{\theta^T f(s, a')}}
$$
where the denominator is a sum over all actions possible in state $s$. The weights are updated using a stochastic gradient ascent rule:
$$
\theta \leftarrow \theta + \alpha \nabla_\theta{J(\theta)} 
$$
where $J(\theta)$ is the expected return from the policy according to the distribution over trajectories induced by the policy. 

%Further, we do not directly learn $\pi$. We learn a surrogate function $w(a \vert s; \theta)$ that assigns a weight to each action in the beam of candidate actions. 

%These weights are normalized using a softmax to obtain a probability distribution from which the agent's next action is sampled. 

The state consists of the predicates in the current description, the candidate objects, and the current classifiers. 
Since both the number of candidate objects and classifiers varies, and the latter is quite large, it is necessary to identify useful features for the task to obtain a vector representation needed by most learning algorithms. In our problem setting, the number of candidate actions available to the agent in a given state is variable. 
Hence we need to create features for state-action pairs, rather than just states. 

\subsection{Features for Policy Learning}

The object retrieval task consists of two parts -- identifying useful queries to improve classifiers, and correctly guessing the image being referred to by a given description. 
The current dialog length is also provided to influence the trade-off between guessing and querying.
%It is reasonable to expect that some features are primarily relevant to predicting the usefulness of queries, some are relevant to estimating whether a guess made in the current state is likely to succeed, and some features can be used to trade-off between the two. 

\subsubsection{Guess-success features}

Let $P_A = \{p_1, p_2, \ldots p_k\}$ be the predicates extracted from the current description. For each predicate $p \in P_A$, we have the estimated F1 of the classifier $C(p)$, and for each object $o$ in the active test set, we have a decision $d(p, o) \in \{-1, 1\}$ from the classifier. We refer to $s(p, o) = d(p, o) * C(p)$ as the score of the classifier of $p$ for object $o$. The following features are used to predict whether the current best guess is likely to be correct:
\begin{itemize}
\setlength\itemsep{0em}
\item Lowest, highest, second highest, and average estimated F1 among classifiers of predicates in $P_A$ -- learned thresholds on these values can be useful to decide whether to trust the guess.
\item Highest score among regions in the active test set, and the differences between this and the second highest, and average scores respectively -- a good guess is expected to have a high score to indicate relevance to the description, and substantial differences would indicate that the guess is discriminative. Similar features are also formed using the unweighted sum of decisions. 
\item An indicator of whether the two most confident classifiers agree on the decision of the top scoring region, which increases the likelihood of its being correct.
%\item Difference between the decision for the top scoring region by the two most confident classifiers, and their respective average decisions to determine whether these classifiers are discriminative. 
\end{itemize}

We compared directly using these features to training a regressor that uses them to predict the probability of a successful guess, and then using this as a higher-level policy feature. We found no difference between the two methods and the results reported directly use these features in the vector provided to the policy learner. 

\subsubsection{Query-evaluation features}

\noindent The following features are expected to be useful in predicting whether it is useful to query for the label of a particular predicate:
\begin{itemize}
\setlength\itemsep{0em}
\item Indicator of whether the predicate is new or already has a classifier -- this allows the policy to decide between strengthening existing classifiers or creating classifiers for novel predicates.
\item Current estimated F1 of the classifier for the predicate -- as there is more to be gained from improving a poor classifier.
\item Fraction of previous dialogs in which the predicate has been used, and the agent's success rate in these -- as there is more to be gained from improving a frequently used predicate but less if the agent already makes enough correct guesses for it.
\item Is the query opportunistic -- as these will not help the current guess.
%\item Fraction of previous dialogs in which the predicate has been used where the agent successfully guessed the image described -- as there is less to be gained from a predicate that already has a classifier good enough to result in correct guesses.
\end{itemize}

Label queries also have an image region specified, and for these we have additional features that use the VGG feature space in which the region is represented for classification:
\begin{itemize}
\setlength\itemsep{0em}
\item Margin of the image region from the hyperplane of the classifier of the predicate -- motivated by uncertainty sampling.
\item Average cosine distance of the image region to others in the dataset -- motivated by density weighting to avoid outliers.
\item Fraction of the $k$-nearest neighbors of the region that are unlabeled for this predicate -- motivated by density weighting to identify a data point that can influence many labels.
\end{itemize}

%\subsection{Common features}

%\noindent The following features are aimed at helping the agent trade-off between guessing and querying: 
%\begin{itemize}
%\item The type of action being evaluated - guess, label query or example query.
%\item If the agent is evaluating a query, is it a predicate present in the current description.
%\item Current dialog length -- to favor shorter dialogs.
%\end{itemize}

\section{Experimental Methodology}

\subsection{Dataset}

The Visual Genome dataset contains a total of 108,077 images with 540,6592 annotated regions. Since objects and attributes are annotated with free-form text rather than from a fixed, pre-defined vocabulary, there is considerable diversity in the language used for annotation. There are 80,908 unique objects annotated and 44,235 attributes. We assume that any objects that partially overlap with a region are present in it, as these are usually used in descriptions. Using the annotations, we can associate a list of objects and attributes relevant to each image region.  We lower-case all annotations, remove special characters and perform stemming to help normalize terms.

\subsection{Sampling dialogs}
\label{ssec:sampling}

We want the agent to learn a policy that is independent of the actual predicates present at policy training and policy test time. 
In order to be able to evaluate this, we divide the set of possible regions into policy training and policy test regions as follows. We select all objects and attributes present in at least 1,000 regions. 
Half of these were randomly assigned to the policy test set. 
All regions that contain one of these objects or attributes are assigned to the policy test set, and the rest to the policy training set. 
Thus regions seen at test time may contain predicates seen during training, but will definitely contain at least one novel predicate. 
Further, the policy training and policy test sets are respectively partitioned into a classifier training and classifier test set using a uniform 60-40 split. 

% Dialogs are then sampled as described in section \ref{sec:task_setup}. 
During policy training, the active training set of each dialog is sampled from the classifier-training subset of the policy-training regions, and the active test set of the dialog is sampled from the classifier-test subset of the policy-training set. 
During policy testing, the active training set of each dialog is sampled from the classifier training subset of the policy test regions, and the active test set of the dialog is sampled from the classifier test subset of the policy test set. 

\subsection{Experiment phases}

For efficiency, we run dialogs in batches, and perform classifier and policy updates at the end of each batch. We use batches of 100 dialogs each. Our experiment runs in 3 phases: 
\begin{itemize}
\item Initialization -- Since learning starting with a random policy can be difficult, we first run batches of dialogs on the policy training set using the static policy from section \ref{ssec:static_policy}, and update the RL policy using states, actions and rewards seen in these dialogs. This ``supervised'' learning phase is used to initialize the RL policy. % TODO: Add citations for this
\item Training -- We run batches of dialogs on the policy training set using the RL policy, starting it without any classifiers. In this phase, the policy is updated using its own experience. 
\item Testing -- We fix the parameters of the RL policy, and run batches of dialogs on the policy test set. During this phase, the agent is again reset to start with no classifiers. We do this to ensure that performance improvements seen at test time are purely from learning a strategy for opportunistic active learning, not from acquiring useful classifiers in the process of learning the policy.  
\end{itemize}

\section{Experimental Results and Analysis}

We initialize the policy with 10 batches of dialogs, and then train on another 10 batches of dialogs, both sampled from the policy training set. Following this, the policy weights are fixed, the agent is reset to start with no classifiers, and we test on 10 batches of dialogs from the policy test set. 
Table \ref{tab:group_ablation} compares the average success rate (fraction of successful dialogs in which the correct object is identified), and average dialog length (average number of system turns) of the best learned policy, and the baseline static policy on the final batch of testing. We also compare the effect of ablating the two main groups of features. The learned agent guesses correctly in a significantly higher fraction of dialogs compared to the static agent, using a significantly lower number of questions per dialog. 

When either the group of guess or query features is ablated, the success rate clearly decreases. While the mean success rate still remains above the baseline, the difference is no longer statistically significant. Further, at the end of the initialization phase, the average dialog length in all three conditions is about the same. In the two ablated conditions, the dialog length does not increase to become close to that of the static policy, which suggests that the agent does not learn that asking more queries improves dialog success. This is expected because the agent is either not able to evaluate the usefulness of queries, or the likelihood of success of a guess. 
However, in the learned policy with all features, the agent is able to identify a benefit in asking queries, and utilizes them to improve its success rate.

It is important to note that it is non-trivial to decide how to trade-off dialog success with dialog length.
This should be decided for any given application by comparing the cost of an error with that of the user time involved in answering queries, and the reward function should be set appropriately based on this.
Ideally, we would like to see an increase in dialog success rate {\it and} a decrease in dialog length, as is the case when comparing the learned and static policies. 
However, depending on the application, it may also be beneficial to see a smaller increase in success rate with a larger decrease in dialog length, as is the case in the ablated conditions.
 
\begin{table}
\begin{tabular}{|c|c|c|}
\hline
Policy & Success rate & Average Dialog Length \\ \hline 
Learned & \textbf{0.44} & \textbf{12.95} \\ 
--Guess & 0.37 & \textbf{6.12} \\ 
--Query & 0.35 & \textbf{6.16} \\ 
Static & 0.29 & 16 \\ \hline 
\end{tabular} 
\caption{Results on dialogs sampled from the policy test set after 10 batches of classifier training. \textit{--Guess} and \textit{--Query} are conditions with the guess and query features, respectively, ablated. Boldface indicates that the difference in that metric with respect to the \textit{Static} policy is statistically significant according to an unpaired Welch t-test with $p < 0.05$.}
\label{tab:group_ablation}
\end{table}

We also explored ablating individual features. We found that the effect of ablating most single features is similar to that of ablating a group of features. 
The mean success rate decreases compared to the full policy with all features. 
It remains better than that of the static policy, but in most cases the difference stops being statistically significant. 
Among features for evaluating the guess, the removal of the difference between the two highest scores in the active test set has a fairly large effect, compared with the value of the highest score. 
This is expected because for retrieval it is sufficient if an object is simply scored higher than the other candidates. 
Further, since classifiers improve over time, the score threshold that indicates a good guess changes, and hence would be difficult to learn. 
An interesting result is that removal of features involving the predictions of the second best classifier has more effect than that of the best classifier. 
This is possibly because when noisy classifiers are in use, support of multiple classifiers is helpful. 
Among query evaluation features, we find, unsurprisingly, that removal of the feature providing the margin of the object in a label query affects performance much more than removal of features such as density and fraction of labeled neighbors, which merely indicate whether the object is an outlier.  
The full results of this experiment are included in the supplementary material. 

Qualitatively, we found that the dialog success rate was higher for both short, and very long dialogs, with a decrease for dialogs of intermediate length.
This suggests that longer dialogs are used to accumulate labels via opportunistic off-topic questions, as opposed to on-topic questions. 
The learned policy still suffers from high variance in dialog length suggesting that trading off task completion against model improvement is a difficult decision to learn. 
We find that the labels collected by the learned policy are more equitably distributed across predicates than the static policy, resulting in a tendency to have fewer classifiers of low estimated F1.
There is relatively little difference in the number of predicates for which classifiers are learned. 
%This suggests that the policy learns to exploit the sampling distribution, which was originally tailored for the baseline, to focus on select predicates, while being able to improve them more uniformly than the distribution directly allows. 
This suggests that the policy learns to focus on a few predicates, as the baseline does, but learn all of these equally well, in contrast to the baseline which has much higher variance in the number of labels collected per predicate. 

\section{Future Work}

It would be interesting to examine how a policy learned using a dataset such as Visual Genome generalizes to a different domain such as images captured by a robot operating in an indoor environment, possibly with some fine-tuning using a smaller in-domain dataset.
The simulation could also potentially be improved using positive-unlabeled learning methods \cite{liu:icml02,li:ijcai03} instead of assuming that an object or attribute not labeled in an image region is not present in the image.
It would also be interesting to compare the effectiveness of the opportunistic active learning framework, as well as the policy learning, across a variety of applications.

\section{Conclusion}

This paper has shown how to formulate an opportunistic active learning problem as a reinforcement learning problem, and learn a policy that can effectively trade-off opportunistic active learning queries against task completion. We evaluated this approach on the task of grounded object retrieval from natural language descriptions and learn a policy that retrieves the correct object in a larger fraction of dialogs than a previously proposed static baseline, while also lowering average dialog length.

\section*{Acknowledgements}
This work is supported by an NSF NRI grant (IIS-1637736).
A portion of this work has taken place in the Learning Agents Research Group (LARG) at UT Austin.
LARG research is supported in part by NSF (CNS-1305287, IIS-1637736, IIS-1651089, IIS-1724157), TxDOT, Intel, Raytheon, and Lockheed Martin.
Peter Stone serves on the Board of Directors of Cogitai, Inc.
The terms of this arrangement have been reviewed and approved by the University of Texas at Austin in accordance with its policy on objectivity in research.

\bibliography{emnlp2018}
\bibliographystyle{acl_natbib}

\appendix 

\section{Supplemental Material}
\label{sec:supplemental}

\subsection{Sampling predicates for label queries}

A large number of perceptual predicates can be used to describe objects. When choosing a predicate whose classifier is to be improved by active learning, the simplest way to choose between predicates is to favor those for which the agent currently has poor classifiers. 
However, if the number of possible predicates is much larger than the number of queries available for learning, it becomes necessary to focus on a small number of predicates, possibly stopping the improvement on a predicate once the classifier for it has been sufficiently improved. 

We use the following distribution to obtain probability weights for predicates as a function of estimated classifier F1. Let $w(p_i)$ be the weight for predicate $p_i$ with estimated F1 $C(p_i)$. Weights start at $w_{min}$ for $C = 0.0$ and increase linearly to $w_{max}$ at some $C_{max} \in (0, 1)$. For $C >  C_{max}$, weights again linearly decrease to $w_{min}$ for $C = 1.0$. That is, for $C <= C_{max}$, 
$$
w(p_i) = \frac{C(p_i)}{C_{max}} (w_{max} - w_{min})
$$
For $C > C_{max}$,
$$
w(p_i) = \frac{1.0 - C(p_i)}{1.0 - C_{max}} (w_{max} - w_{min})
$$
The weights are then normalized to obtain a probability distribution. A beam of label queries can then be sampled from this.

\subsection{Results of Individual Feature Ablation}

Table \ref{tab:individual_ablation} contains the results of ablation of individual features. We use the notation $P_A = \{p_1, p_2, \ldots p_k\}$ for the predicates extracted from the current description. For each predicate $p \in P_A$, we have the estimated F1 of the classifier $C(p)$, and for each object $o$ in the active test set, we have a decision $d(p, o) \in \{-1, 1\}$ from the classifier. We refer to $s(p, o) = d(p, o) * C(p)$ as the score of the classifier of $p$ for object $o$. The best classifier for the current interaction is the one with maximum estimated F1, that is, the classifier for $p_{best} = argmax_{p \in P_A}{C(p)}$. The second best classifier is $p_{sec} = argmax_{p \in P_A-{p_{best}}}{C(p)}$

% We find that most features result in the mean success rate remaining better than the baseline, but the difference stops being statistically significant. 

\begin{table}[H]
\begin{tabular}{|L{4cm}|C{1.5cm}|C{1.2cm}|}
\hline
Feature Ablated & Success rate & Average Dialog Length \\ \hline \hline 
None & \textbf{0.44} & \textbf{12.95} \\ \hline 
Number of system turns used - normalized & 0.41 & \textbf{3.8} \\ \hline 
Density of object in label query & 0.4 & \textbf{12.89} \\\hline 
Fraction of previous dialogs using predicate in query that have succeeded & 0.39 & \textbf{5.46} \\ \hline 
Score (normalized) of top region & 0.39 & \textbf{6.3} \\ \hline 
Fraction of k nearest neighbors of the object in label query, which are unlabeled & 0.39 & \textbf{10.41} \\ \hline 
Indicator for guess action & 0.38 & \textbf{7.21} \\ \hline 
Minimum value of $C(p)$ for $p \in P_A$ & 0.37 & \textbf{6.37} \\ \hline 
Decision of $p_{sec}$ for object with highest score & 0.37 & \textbf{11.21} \\ \hline 
Difference between decision of $p_{best}$ for object with highest score, and the average of its decisions for objects in the active test set & 0.36 & \textbf{2.78} \\ \hline 
\end{tabular} 
\end{table}

\begin{table}[H]
\begin{tabular}{|L{3.5cm}|C{1.5cm}|C{1.5cm}|}
\hline
Feature Ablated & Success rate & Average Dialog Length \\ \hline  \hline 
Indicator of whether the question is on-topic & 0.36 & \textbf{5.25} \\ \hline 
Is decision of $p_{best}$ same for objects with top two scores & 0.36 & \textbf{5.32} \\ \hline 
Indicate of whether the predicate in the query has a classifier & 0.36 & \textbf{13.97} \\ \hline
Frequency of use of the predicate in query - normalized & 0.35 & \textbf{4.48} \\ \hline 
Indicator for the action of asking a positive example & 0.35 & \textbf{5.36} \\ \hline 
Second highest value of $C(p)$ for $p \in P_A$ & 0.35 & \textbf{5.9} \\ \hline 
Current estimated F1 for classifier of the predicate in query & 0.35 & \textbf{6.53} \\ \hline 
Decision of $p_{best}$ for object with highest score & 0.34 & \textbf{3.85} \\ \hline 
Indicator for the action of asking a label & 0.34 & \textbf{7.05} \\ \hline 
Average value of $C(p)$ for $p \in P_A$ & 0.34 & \textbf{7.63} \\ \hline 
Margin of object in label query & 0.34 & \textbf{8.08} \\ \hline 
Maximum value of $C(p)$ for $p \in P_A$ & 0.33 & \textbf{6.31} \\ \hline 
Difference between decision of $p_{sec}$ for object with highest score, and the average of its decisions for objects in the active test set & 0.33 & \textbf{8.84} \\ \hline 
Difference between top two scores in the active test set & 0.32 & \textbf{8.05} \\ \hline 
Is decision of $p_{best}$ same for objects with top two scores & 0.32 & \textbf{10.01} \\ \hline 
Difference between top score and average score in the active test set & 0.31 & \textbf{7.18} \\ \hline 
Baseline & 0.29 & 16 \\ \hline  
\end{tabular} 
\caption{Results of individual feature ablation. Boldface indicates that the difference in that metric with respect to \textit{Static} is statistically significant according to an unpaired Welch t-test with $p < 0.05$.}
\label{tab:individual_ablation}
\end{table}

\end{document}